\documentclass[letterpaper, 10pt, conference, hidelinks]{ieeeconf}

\usepackage{etoolbox}
\makeatletter
\patchcmd{\@makecaption}
  {\scshape}
  {}
  {}
  {}
\makeatother

\usepackage{xfp} % 
\usepackage{multirow}

\IEEEoverridecommandlockouts                              

\usepackage{graphicx} % for pdf, bitmapped graphics files
\usepackage{subcaption}
\usepackage{amsmath} % assumes amsmath package installed
\usepackage{amssymb}  % assumes amsmath package installed
\usepackage{url}
\usepackage{adjustbox}
\usepackage[draft]{pdfcomment} % https://ctan.org/pkg/pdfcomment?lang=en 
\usepackage[hyperfirst=false,acronym,sort=none,shortcuts,nopostdot,style=super,nonumberlist,toc,nogroupskip]{glossaries}
\usepackage[dvipsnames]{xcolor} % used for colored acronyms
\usepackage{xspace}

\glsdisablehyper % to disable hyperlinks on all acronyms

\definecolor{AcColor}{rgb}{.15,0,0}

\newcommand{\accolor}[1]{\textcolor{AcColor}{#1}} % subsequent uses
\newacronymstyle{myacro}
{%
  \GlsUseAcrEntryDispStyle{long-short}%
}%
{%
  \GlsUseAcrStyleDefs{long-short}%
}

\setacronymstyle{myacro}

\newcommand*{\tip}[1]{%  define our acronym command,  make it short since we use it a lot, use * for so that it is only a 'short' command
    \ifglsused{#1}{% if we used it already, then put pdftooltip
      {\pdftooltip{\accolor{\glsentryshort{#1}}}{\glsentrydesc{#1}}}%
    }{%
      \gls{#1}% otherwise put the normal gls
    }%
}%

\let\xx\tip
\let\xxx\tips
\newacronym{adc}{ADC}{Analog to Digital Converter}
\newacronym{admm}{ADMM}{Alternating Direction Method of Multipliers}
\newacronym{asic}{ASIC}{Application Specific Integrated Circuit}
\newacronym{cem}{CEM}{Cross-Entropy Method}
\newacronym{cots}{COTS}{Commodity Off-The-Shelf}
\newacronym{cpu}{CPU}{Central Processing Unit}
\newacronym{ddp}{DDP}{Differential Dynamic Programming}
\newacronym{dnn}{DNN}{Deep Neural Network}
\newacronym{dof}{DOF}{Degree Of Freedom}
\newacronym{dram}{DRAM}{Dynamic RAM}
\newacronym{fpga}{FPGA}{Field Programmable Gate Array}
\newacronym{gpu}{GPU}{Graphics Processing Unit}
\newacronym{gru}{GRU}{Gated Recurent Unit}
\newacronym{hil}{HIL}{Hardware In the Loop}
\newacronym{hls}{HLS}{High Level Synthesis}
\newacronym{ip}{IP}{Interior Point}
\newacronym{ipb}{IP block}{Intellectual Property Block}
\newacronym{lstm}{LSTM}{Long Short Term Memory}
\newacronym{ltv}{LTV}{Linear Time Varying}
\newacronym{mac}{MAC}{Multiply-Accumulate}
\newacronym{mae}{MAE}{Median Angular Error}
\newacronym{mlp}{MLP}{Multilayer Perceptron}
\newacronym{mee}{MEE}{Median Endpoint Error}
\newacronym{mpc}{MPC}{Model Predictive Control}
\newacronym{mpcer}{MPC}{Model Predictive Controller}
\newacronym{mppi}{MPPI}{Model Predictive Path Integral}
\newacronym{nc}{NC}{Neural Controller}
\newacronym{nmpc}{NMPC}{Nonlinear Model Predictive Control}
\newacronym{nn}{NN}{Neural Network}
\newacronym{npc}{NPC}{Neural Predictive Control}
\newacronym{npu}{NPU}{Neural Processing Unit}
\newacronym{ode}{ODE}{Ordinary Differential Equation}
\newacronym{pcb}{PCB}{Printed Circuit Board}
\newacronym{pd}{PD}{Proportional Derivative}
\newacronym{pid}{PID}{Proportional Integral Derivative}
\newacronym{pl}{PL}{Programmable Logic}
\newacronym{pp}{PP}{Pure Pursuit}
\newacronym{ps}{PS}{Processing System}
\newacronym{pso}{PSO}{Particle Swarm Optimization}
\newacronym{pwm}{PWM}{Pulse Width Modulation}
\newacronym{qp}{QP}{Quadratic Programming}
\newacronym{rl}{RL}{Reinforcement Learning}
\newacronym{rma}{RMA}{Rapid Motor Adaptation}
\newacronym{rnn}{RNN}{Recurrent Neural Network}
\newacronym{roa}{ROA}{Regularized Online Adaptation}
\newacronym{ros}{ROS}{Robot Operating System}
\newacronym{rpgd}{RPGD}{Resampling Parallel Gradient Descent}
\newacronym{slp}{SLP}{Single Layer Perceptron}
\newacronym{sm}{SM}{Supplementary Material}
\newacronym[description={System on Chip; FPGA with embedded programmable processor}]{soc}{SoC}{System on Chip}
\newacronym{sqp}{SQP}{Sequential Quadratic Programming}
\newacronym{sram}{SRAM}{Static RAM}
\newacronym{usb}{USB}{Universal Serial Bus}
\newacronym{vga}{VGA}{Video Graphics Adaptor}
\newacronym{uart}{UART}{Universal Asynchronous Receiver/Transmitter}
\newacronym{xla}{XLA}{Accelerated Linear Algebra}

\usepackage[abbreviations]{foreign} % https://tex.stackexchange.com/questions/15009/macros-for-common-abbreviations
\usepackage{makecell} % line breaks in table cells https://tex.stackexchange.com/questions/2441/how-to-add-a-forced-line-break-inside-a-table-cell/19678
\usepackage[capitalise]{cleveref} % allows use of \cref for figs, tables, sections etc.

\usepackage[leftcaption]{sidecap} % https://www.ctan.org/pkg/sidecap 

\usepackage{hyperref}  %hyperref still needs to be put at the end!

\usepackage[style=ieee, backend=biber,maxnames=6]{biblatex} % https://tex.stackexchange.com/questions/424774/print-url-only-if-doi-not-present
\DeclareSourcemap{
  \maps[datatype=bibtex]{
    \map[overwrite]{
      \step[fieldsource=doi, final]
      \step[fieldset=url, null]
      \step[fieldset=eprint, null]
    }  
  }
}
\usepackage{gensymb} % \degree symbol and others

\usepackage{xfp} % 
\title{\LARGE \bf
FPGA 
Hardware Neural Control of CartPole and F1TENTH Race Car$^{0}$
}

\author{Marcin~Paluch$^{\ast,\ddagger}$, Florian Bolli$^{\ast,\ddagger}$, Xiang Deng$^{\ast}$, Antonio Rios Navarro$^{\dagger}$,\\ Chang Gao$^{\mathsection}$, and Tobi~Delbruck$^{\ast}$%
\thanks{$0$ This work has been submitted to the IEEE for possible publication. Copyright may be transferred without notice, after which this version may no longer be accessible.}
\thanks{$\ddagger$ Contributed equally to this work by doing the bulk of development and experiments.}
\thanks{$^{\ast}$ Sensors Group, Inst. of Neuroinformatics, UZH-ETH Zurich; \url{https://sensors.ini.ch}. }%
\thanks{$^{\dagger}$ U. of Seville; \url{https://investigacion.us.es/sisius/sis_showpub.php?idpers=20628}. }%
\thanks{$^{\mathsection}$ TU Delft; \url{https://www.tudemi.com/}. }%
}

\def\urlhls4ml{\url{https://github.com/fastmachinelearning/hls4ml}}

\def\urlcartpole{\url{https://aliexpi.com/3fGL}}
\def\urlzybo{\url{https://digilent.com/reference/programmable-logic/zybo-z7/start}}

\def\urlEKF{\url{http://wiki.ros.org/robot_pose_ekf}}
\def\urlAMCL{\url{https://wiki.ros.org/amcl}}
\def\urlSLAM{\url{https://google-cartographer-ros.readthedocs.io/en/latest/}}

\def\urlgithubpaper{\url{https://github.com/SensorsINI/Neural-Control-Tools}}

\begin{document}

\maketitle

\thispagestyle{empty}
\pagestyle{empty}

%%%%%%%%%%%%%%%%%%%%%%%%%%%%%%%%%%%%%%%%%%%%%%%%%%%%%%%%%%%%%%%%%%%%%%%%%%%%%%%%
\begin{abstract}

Nonlinear model predictive control (NMPC) has proven to be an effective control method, but it is expensive to compute. 
This work demonstrates the use of hardware FPGA neural network controllers trained to imitate NMPC with supervised learning. We use these Neural Controllers (NCs) implemented on inexpensive embedded FPGA hardware for high frequency control on physical cartpole and F1TENTH race car. 
Our results show that the NCs match the control performance of the NMPCs in simulation and outperform it in reality, due to the faster control rate that is afforded by the quick FPGA NC inference. We demonstrate kHz control rates for a physical cartpole and offloading control to the FPGA hardware on the F1TENTH car. Code and hardware implementation for this paper are available at \urlgithubpaper.

\end{abstract}


%%%%%%%%%%%%%%%%%%%%%%%%%%%%%%%%%%%%%%%%%%%%%%%%%%%%%%%%%%%%%%%%%%%%%%%%%%%%%%%%
\section{INTRODUCTION}
\label{sec:introduction}

The idea of using \xxx{nn} to model system dynamics for predictive nonlinear control dates to the 1990s%
% where it was termed as \xx{npc}
~\cite{Bhat1990-nn-for-dc-early,Narendra1990-id-and-control-of-dyn-sys-using-nns-11k-cites,Saint-Donat1991-nn-mpc-ph-cstr,Draeger1995-mpc-using-nns-ph-cstr}.
Vast improvements in \xxx{dnn} and \xx{rl} have  resulted in impressive demonstrations of direct neural control of difficult nonlinear robotic control problems, e.g. for drone racing~\cite{Kaufmann2023-rpg-drone-racing-nature} and quadruped navigation ~\cite{Miki2022-hutter-rsl-rl-quadroped,Lee2020-rsl-quadroped-locomotion-rl}. Their \xx{dnn} controllers are trained to infer optimal control actions from the current state and target goals. 
In this paper, we use the term \xx{nc} to refer to neural networks that directly generate control actions from state and target goals.

An \xx{nc} maps from a state (or set of past states) and a set of goals to a control action by computing a \xx{dnn}. The work reported in this paper shows that inexpensive \xx{soc} \xxx{fpga} can completely offload the control computation with \xx{mlp} 
controllers at high control frequency.

\begin{figure}[t]
  \centering
      \includegraphics[width=.7\columnwidth]{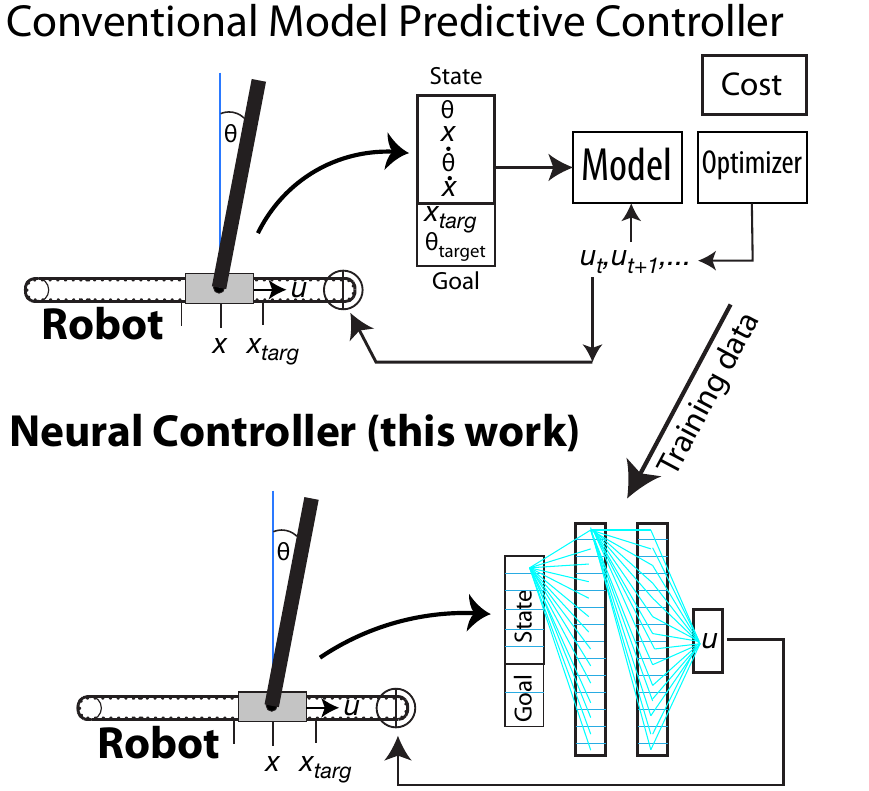}
      \caption{A lightweight hardware accelerated neural controller replaces a conventional expensive \xx{nmpc} controller. In this work, the \xx{nc} is trained by data collected from the \xx{nmpc} in simulation.}
  \label{fig:fig1-concept}
\end{figure}

\cref{fig:fig1-concept} illustrates how we propose to replace a conventional \xx{nmpc} controller with a lightweight \xx{nc} that has been trained in simulation 
% (augmented with real data) 
to imitate an expensive \xx{nmpc}.
%In contrast to ~\cite{Song2023-mpc-vs-rl-drone-racing}, 
The \xx{nmpc} objective used to gather training data is crafted to achieve a desired goal over its rollout horizon. Like with \xx{nmpc}, implicit feedback control occurs at every timestep.

Our approach follows previous work on hardware accelerated \xx{nmpc}~\cite{Wang2022-ann-mpc,Xu2022-pso-mpc-fpga,Saint-Donat1991-nn-mpc-ph-cstr,Gao2020-ampro-edgedrnn} but is distinguished by being a pure \xx{nc} that infers the optimal control from what it has learned in simulation, rather than solving a constrained nonlinear optimization problem at each time step.  The \xx{dnn} requires many mathematical operations per time step, but is accelerated using an open framework for hardware \xxx{nn} developed by the particle physics community for \xxx{mlp}~\cite{Fahim2021-hls4ml}.
% and by neuromorphic engineers for \xxx{rnn}~\cite{Gao2020-edgedrnn}. 
This approach enables quicker control than with \xx{nmpc}, at least in some scenarios leading to smoother control and faster corrections. 

The complete controller is implemented on an entry-level USB-powered \xx{fpga} \xx{soc} board that costs about \$300 and burns less than 3W. It runs the multilayered \xxx{nn} with thousands of parameters in microseconds by massive parallelization of the \xx{mac} operations. This forms the complete standalone controller for our cartpole,
and it is a hardware node in our more complex F1TENTH race car autonomous driving system stack.

\subsection{CONTRIBUTIONS AND NOVELTY}
\label{subsec:contributions}

\begin{enumerate}
    \item We contribute an open source framework\footnote{\urlgithubpaper} for the training and implementation of \tip{nmpc} neural imitators on \tip{fpga} accelerated \xxx{nc}.
    \item We contribute concrete implementations of hardware \xxx{nc}  for cartpole and F1TENTH, demonstrating their equivalent control performance to \xx{nmpc} with less power and latency.
    The hardware cartpole \xx{nc}  matches the \xx{nmpc} control performance but allows controlling the cartpole at 1\,kHz.
    The  F1TENTH  \xx{nc} nearly matches the control performance (laptime) of the \xx{nmpc} controller in simulation. On the physical car---where we cannot compute the original \xx{nmpc} sufficiently quickly---it offloads all control computation to the \xx{soc} \xx{fpga}. We show that it achieves more precise tracking and  laps that are completed 20\% faster than an optimized pure pursuit controller.

\end{enumerate}

\section{RELATED WORK}
\label{sec:related_work}

There is considerable literature on hardware-accelerated solution of \xx{nmpc} optimization: \xx{qp} methods \cite{jerez2011fpga},
\xx{ip}~\cite{He2005-ip-mpc-on-chip-simulation,Ling2006-fpga-qp-fpga-aircraft-elevator,liu2014fpga,jerez2010fpga}, 
fast gradient and \xx{admm}~\cite{Jerez2013-fpga-gradient-method-atomic-force-microscope,dang2015embedded, zhang2017embedded}, \xx{pso}~\cite{Xu2022-pso-mpc-fpga}, and others~\cite{Ravera2023-mads-mpc-fpga}.

\xx{qp} solves convex optimization, especially quadratic functions. \xx{sqp} is an iterative process that linearizes a nonlinear function and turns it into a sequence of \xx{qp} optimizations. \xx{sqp}, the dominant solver used for \xx{nmpc}, is difficult to parallelize in hardware and is also dependent on correctly adjusting many parameters. \xxx{nc} are a simpler solution that leverages offline \xx{nmpc} for simpler implementation of nonlinear optimal control.

We could find only few reports of prior hardware \xxx{nc}. Diadati~\cite{Diodati2022-pd-fpga-imitation} mimicked a \xx{pd} power supply controller with a custom \xx{fpga} implementation of a tiny \xx{slp} with only 5 hidden units.
Wang~\cite{Wang2022-ann-mpc,Wang2022-nn-ampc} proposed ANN-MPC (analog neural network MPC) to control a power converter by training a similar 5-hidden-unit \xx{slp} with transfer learning to mimic the original \xx{nmpc}.
Dong~\cite{Dong2023-standoff-tracking-nmpc-nn-fpga} considered simulated drone standoff tracking (flying in circles) and implemented a \xx{nc} using a 2-layer 100 units per layer \xx{mlp} on a \$1,500 \xx{fpga}. 
The most closely related work is \cite{Gao2020-ampro-edgedrnn},  which demonstrated the use of an open source 
\xx{rnn}~\cite{Gao2020-edgedrnn} trained to imitate a PD leg prosthetic controller for the knee and ankle joint torques.

\section{METHODS}
\label{sec:methods}

\subsection{Neural Controller Accelerator}
\label{subsec:nc_accelerator}

Our \xxx{nc} are implemented on a 
Zynq-7020 \xx{soc} hosted on a Zybo Z7-20 development board\footnote{\urlzybo}. 
% This entry-level board costs about \$300.
This \xx{soc} combines a \xx{ps} and \xx{pl} on one chip. The Zynq-7020 includes a 12-bit \xx{adc} and other peripherals like full-speed USB serial port. Zybos's 6-pin PMOD connectors simplify connections to external hardware and we use them to communicate with our cartpole robot. The \xx{ps} (667 MHz dual core Cortex-A9 processor) runs a bare metal C program that interfaces the cartpole robot or theF1TENTH car computer to the \xx{nc} implemented on the \xx{fpga} fabric.
We designed peripheral \xx{pl} components (\xx{pwm}, \xx{adc}, Encoder, Median Filter) with Xilinx Vivado \xx{hls}.

We used the \textsf{hls4ml} framework\footnote{{\urlhls4ml}} to synthesize the \xxx{mlp} for our \xxx{nc}. 
We trained the \xxx{mlp} using QKeras\footnote{\url{https://github.com/google/qkeras}}, a quantized version of Keras. In QKeras, layers are specified as in Keras, but the layer specifcation includes the quantization of weights and activations. We provide these as QM.N fixed point numbers with M total bits and N-bit integer part including sign bit.

\cref{tab:nn_params} lists the \xx{mlp} parameters used for the cartpole and F1TENTH \xxx{nc}, the \xx{pl} utilization, and the compute latency.
We were mainly constrained by the number of LUTs and DSPs. The demand for DSPs can be easily traded against latency, but the usage of LUTs correlates strongly with the network weights and this forced us to experiment with smaller networks, quantization-aware training, and pruning. By rewarding weight pruning in training, we achieved a sparsity of 80\% for the F1TENTH \xx{nc}. Since zero weights do not consume LUTs, it makes the implementation of this larger network possible.
We were constrained by development time and are aware that more optimization (particularly for weights) would be beneficial.
Both \xxx{nc} complete an inference step in less than 4\,us; for the F1TENTH \xx{nc} the \xxx{mac} are computed by parallelism at about 4.5\,GOp/s at a clock frequency of only 25\,MHz.

\begin{table}
    \centering
    \begin{tabular}{|r|l|}
    \hline
\multicolumn{2}{|l|}{\textbf{Cartpole}} \\
        Arch. &  \xx{mlp} 7-32-32-1\\
        \# Params & 1,345\\
        Input Quant. & Q12.2 \\
        Act. Fnc. &  tanh \\
        Weight Quant.  & Q14.4  \\
        Act. Quant. & Q12.1  \\
        Intermediate results & up to 18 bits \\
        Utilization (LUT,FF,BRAM,DSP) & 50\%, 22\%, 11\%, 18\% \\
        Compute latency (Vivado estim.) & 91 cycles, 3.64\,$\mu s$\\
        Power consum. (Vivado estim.) & 12\,mW\\
            \hline
\multicolumn{2}{|l|}{\textbf{F1TENTH}} \\
        Arch. &   \xx{mlp} 64-64-64-2 \\
        \# Params & 16,778 (incl. $\approx$80\% zeros) \\
        Input Quant. & Q16.4 \\
        Act. Fnc. & tanh  \\
        Weight Quant. &  Q16.4   \\
        Act. Quant. & Q16.4 \\
        Intermediate results & up to 16 bits \\
        Utilization (LUT,FF,BRAM,DSP) & 64\%, 28\%, 16\%, 24\%\\
        Compute latency (Vivado estim.) &  93 cycles, 3.72\,$\mu s$\\
        Power consum. (Vivado estim.) & 13\,mW\\
        \hline
    \end{tabular}
    \caption{\xx{nc} parameters, \xx{pl} utilizations, and  latencies.}
    \label{tab:nn_params}
\end{table}

\subsection{Non-linear MPC}
\label{subsec:nonlinear_mpc}

\xx{nmpc} computes a feedback control input $u_k = \mu(\hat{x}_k)$ by minimizing cost $\mathrm{J}$ over a finite horizon of $N$ steps. Let $\bar{u}_{k:k+N-1}$ denote an input plan along the MPC horizon of $N$ timesteps starting from control iteration $k$, and $\hat{x}_{k:k+N}$ the resulting state trajectory according to the approximated model $\hat{\mathrm{f}}$ of system $\mathrm{f}$. Then, the solution control input plan $\nu_{k:k+N-1}$ is
\begin{equation} \label{eq:1}
    \nu_{k:k+N-1} = \underset{\bar{u}_{k:k+N-1} \in \mathcal{U}^{N}}{\mathrm{argmin}}{ \bigg[ \mathrm{J}(\hat{x}_{k:k+N}, \bar{u}_{k:k+N-1})} \bigg]
\end{equation}
subject to
\begin{equation*}
    \begin{gathered}
        \hat{x}_{k+i+1} = \hat{\mathrm{f}}(\hat{x}_{k+i}, \bar{u}_{k+i}) \\
        \hat{x}_{k+i} \in \mathcal{X}, \quad \bar{u}_{k+i} \in \mathcal{U}, \quad i = 0, \ldots, N-1
    \end{gathered}
\end{equation*}
The control taken for the next time step is then $\mu(\hat{x}_k) = \nu_k$.

The cost function $\mathrm{J}$ is crafted to obtain a desired objective. Its terms are usually quadratic and positive definite, but it can also contain extremely nonlinear indicator function terms that specify constraints, such as motor saturation, obstacles, and track edges. 

To solve the optimization problem in~\eqref{eq:1} we used the \xx{rpgd} solver from~\cite{Heetmeyer2023-marcin-rpgd-ini}. We used a desktop PC with Intel Core i7-4790K CPU (4GHz, total power $\approx$150 W).

\subsection{Cartpole}
\label{subsec:methods_cartpole}

\begin{figure}[htb]
  \centering
      \includegraphics[width=.8\columnwidth]{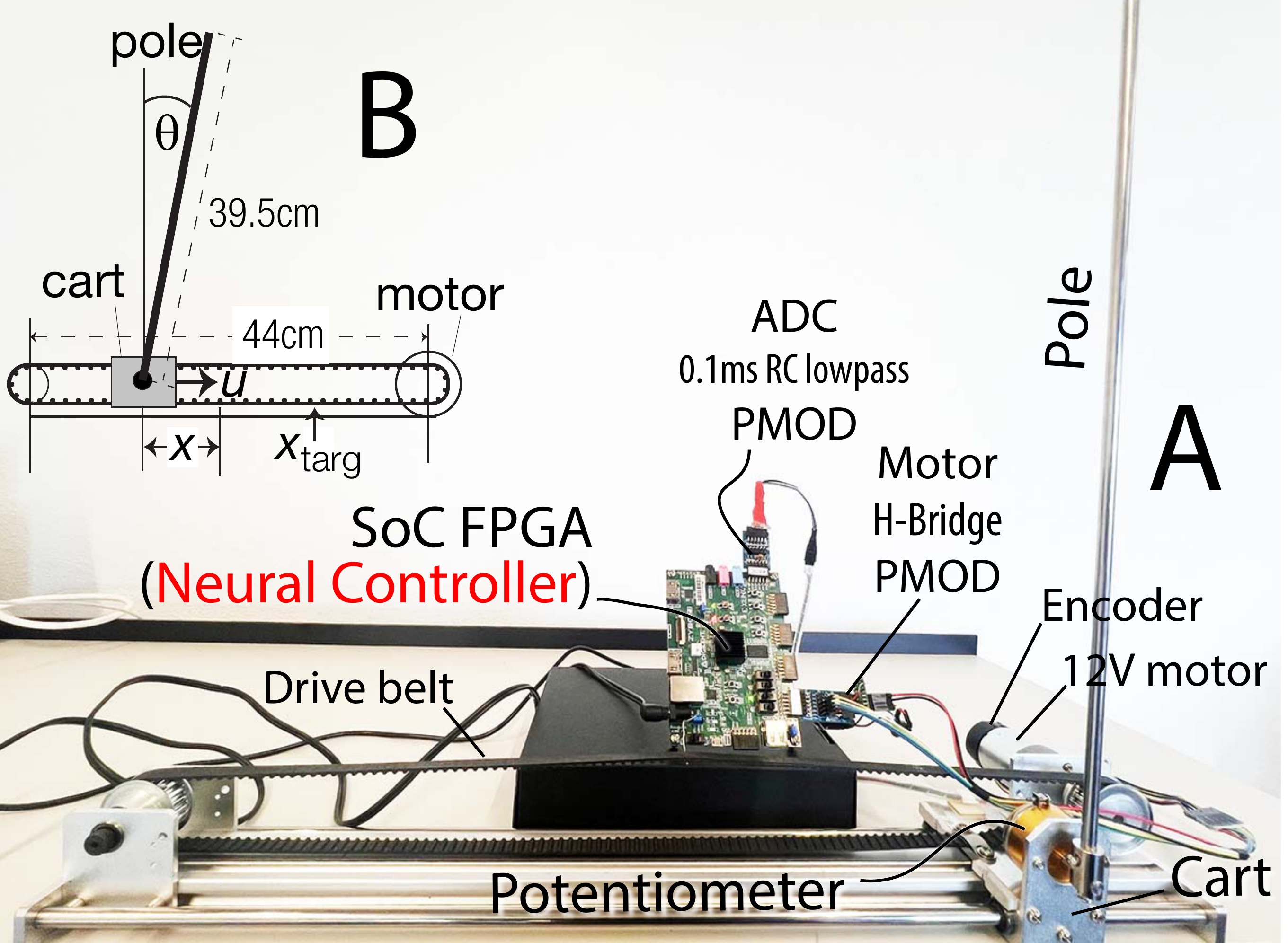}
      \caption{\textbf{A:} Physical cartpole and neural controller hardware. \textbf{B:} Cartpole state variables}
  \label{fig:physical-cartpole-setup}
\end{figure}

{\cref{fig:physical-cartpole-setup}A} shows the physical cartpole\footnote{{\urlcartpole}: Linear Inverted Pendulum} together with the \xx{soc} \xx{fpga} board that runs the \xx{nc}. The \xx{soc} includes a 12-bit \xx{adc} which we run with sampling rate about 350kHz with an additional analog RC filter ($\tau$ = RC = 0.1\,ms) and a rolling 64-sample window median filter on FPGA. The cart motor is controlled by \xx{pwm} \xx{fpga} output that drives the motor H-bridge (TB6612FNG) with 10\,kHz pulses. The motor encoder has 1200 counts per revolution of the gearbox's output shaft, resulting in 118.8 counts per centimeter. The sensor quantization noise is a major obstacle to demonstrate high frequency control - it amplifies while calculating derivatives over short time windows. Hence also for 1\,kHz \xx{nc} we average cart velocity and pole's angular velocity resulting in additional 5\,ms delay for these state components.

\subsubsection{Cartpole Dynamics}
\label{subsubsec:cartpole_dynamics}

{\cref{fig:physical-cartpole-setup}B} illustrates the cartpole state variables. The goal of the cartpole is to swing up the pole and then hold $\theta=0\degree$ (or $\theta=180\degree$) while controlling the cart to a target cart position $x_\text{targ}$ using only horizontal cart acceleration $u$, as quickly as possible, with the least pole motion, cart position error, motor power, and without hitting track boundaries.
The cartpole is underactuated, because it must control pole angle and cart position using a single motor. Cartpole maneuvers require a nonlinear controller,because of the pole dynamics and the constraints on cart position and motor saturation.

\subsubsection{\xx{nc} architecture}
\label{subsubsec:cartpole_nc_architecture}
The \xx{nc} takes as the input state --- sine and cosine of the pole angle, angular velocity, cart position and velocity --- and target --- cart target position and pole target equilibrium --- either up and down. While stabilizing pole down is an easier task, it helps quickly reset the cartpole between experiments and demonstrates that the \xx{nc} can learn two discrete modes of control (with completely opposite signs of small-angle error correction gain), gated by the binary $\theta_\text{targ}$ input.
 The network outputs normalized motor command. It has 2 hidden layers, 32 units each.

\subsubsection{NC data collection and training}
\label{subsubsec:cartpole_data_collection}

We simulated the cartpole using the differential equations of Green~\cite{greenEquationsMotionCart2020}.
These equations are integrated using the standard Euler method with intermediate time steps of 2\,ms. The cart acceleration is calculated by \xx{nmpc} and updated every 20\,ms, both in simulation and in our experiments with physical setup.  We used \xx{rpgd} \cite{Heetmeyer2023-marcin-rpgd-ini} as the optimizer. The controller ran at 50\,Hz with a horizon of 35 time steps (0.7s). 
The cost function contains (weights in brackets) the square penalties on the two minus cosine of the angle (200), angular velocity (13), normalized control signal magnitude (10) and approach to track boundary (active on the extreme 15\% of the track, normalized by track length, 100 000) and linear cost on distance to target position (normalized by track length, 80).
During the simulated experiments, we change the target position, stabilize at the up and down equilibria, and add known control noise to collect a diverse dataset. We collected 1h of simulated data. After data collection we applied state quantization simulating sensor resolution and mixed the dataset with its own copy with 20ms shifted angle and position derivatives to robustify \xxx{nc} against delays.
% coming from various configuration of derivative averaging.

\subsection{F1TENTH car}
\label{subsec:methods_f1tenth_car}

The F1TENTH car is a 1/10 scale autonomous race car. Our goal for this paper is to race it around a track reliably and precisely with a minimum lap time. 

For both simulated and physical experiments, we used a commercial indoor RC race track where the extracted map formed the basis for simulations.
This racetrack is sticky, with an estimated friction coefficient of about 1.1.
An amateur (one of us) can achieve about 24\,s lap times.
The desired race line (minimum curvature line around the given track with a safety margin) was calculated offline with \cite{global_trajectory_optimizer} and saved as a set of waypoints. 
Each waypoint is represented by 3 values: the position ($\text{position}_x$, $\text{position}_y$) and the suggested longitudinal velocity $v_x$.
To control the overall speed of the car, 
we use a speed factor $F$ that multiplies $v_x$ at each waypoint. 
$F$ allows us to easily adjust the car's speed to balance speed and safety.

\subsubsection{Simulated F1TENTH Car}
We used a simulated environment for the F1TENTH car called F1TENTH Gym~\cite{okelly2020f1tenthGym}. The car dynamics are based on the Single-Track car model of the CommonRoads Vehicle Models \cite{commonRoad2020althoff}. For every experiment, the car is initialized at a random position on the minimum-curvature race line, pointing along the race line. To model an estimated 80\,ms delay in perception and actuation, we implemented a control delay from state measurement to actuation in simulation.

\subsubsection{Physical F1TENTH Car}
\label{subsubsec:physical_f1tenth_car}
\begin{figure}[htb]
  \centering
      \includegraphics[width=.8\columnwidth]{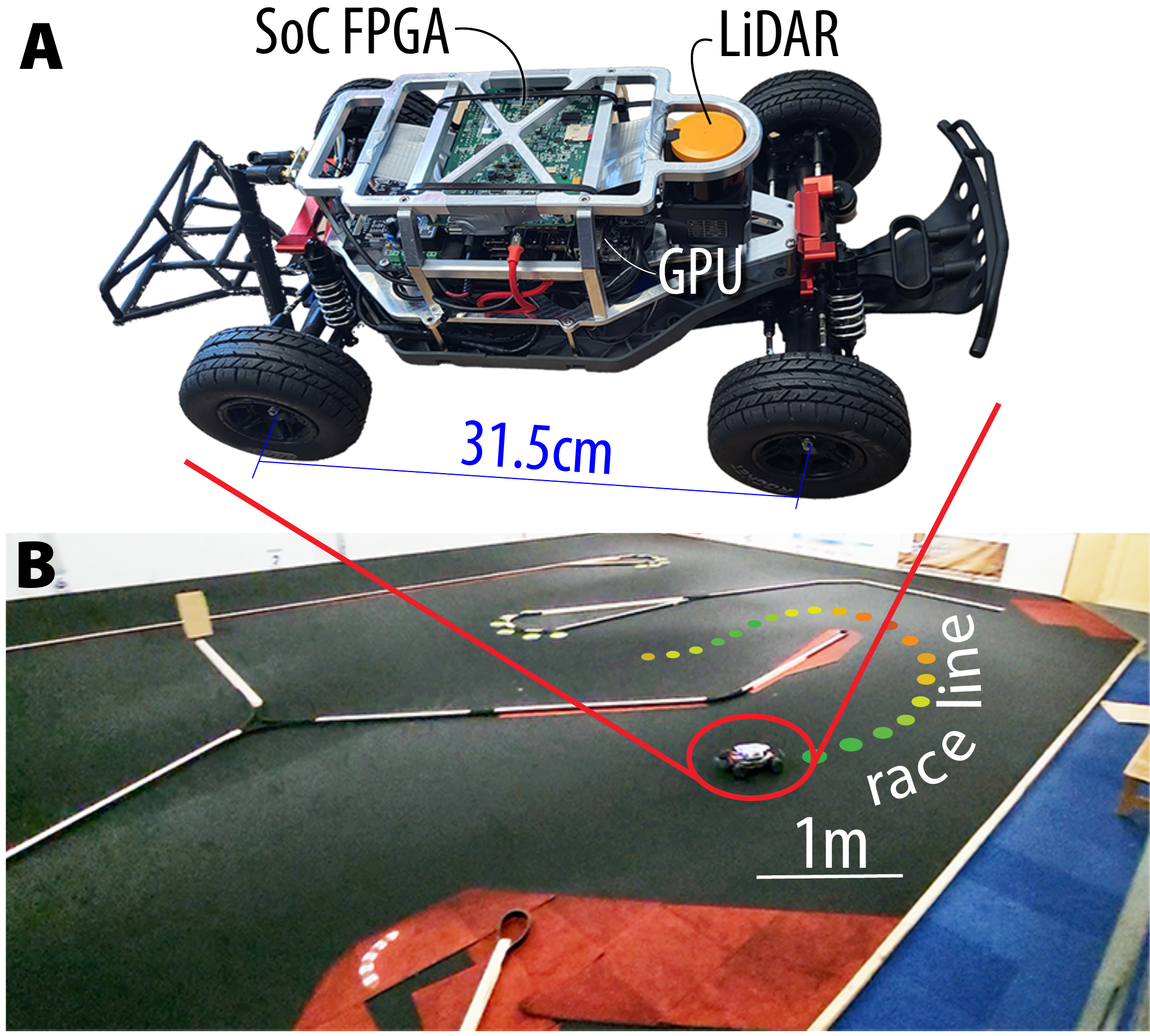}
      \caption{\textbf{A:} F1TENTH car. \textbf{B:} Racetrack (RCA2) including an example of the 20 next waypoints of the race line. The suggested speed is color coded (green: faster,  red: slower)}
  \label{fig:f1tenth_car_track}
\end{figure}

\cref{fig:f1tenth_car_track} shows the physical F1TENTH car and the racetrack. The car is a 1/10 scale Traxxas Slash RC car based on the standard F1TENTH car\footnote{\url{https://f1tenth.org/build.html}}, carrying an NVIDIA Jetson TX2 as the main computer, a Hokuyo LiDAR (10m, 40Hz), a Bosch BNO-6 IMU and a VESC MK6 (electronic speed controller for the brushless motor). We added the \xx{nc} Zybo Z7 \xx{soc} \xx{fpga} of \cref{subsec:nc_accelerator}. 

The state estimation stack consists of an extended Kalman Filter\footnote{{\urlEKF}}, which fuses the motor encoder information with the IMU measurement to obtain odometry information. The odometry position estimation is accurate for a short time window, but it drifts over time due to the differential nature of the IMU. To correct the drifting odometry, the AMCL\footnote{{\urlAMCL}} particle filter is used to estimate the car's position such that the LiDAR scans match the known environment. The map was recorded with the ROS-Cartographer\footnote{{\urlSLAM}} SLAM algorithm.

The car's linear and angular velocity $v_x$ and $\omega_z$ are obtained directly from the Kalman Filter. The steering angle $\theta$ is calculated by the steering control $u_\text{steering}$ and the servo model introduced in \cite{okelly2020f1tenthGym}. The physical car's slipping angle $\beta$ is approximated as $0$, which is justified by the observation that the car rolls over before slipping due to the track's high friction.

\subsubsection{NC data collection and training}
\label{subsubsec:f1tenth_data_collection}
To train our \xx{nc}, we used the \xx{nmpc} controller to collect state-action pairs from the RCA1 track in \cref{fig:f1tenth_car_track}A.
We used \xx{rpgd} \cite{Heetmeyer2023-marcin-rpgd-ini} as the optimizer for our simulated \xx{nmpc} control experiments. The controller ran at 50\,Hz with a horizon of 20 time steps or 0.4s. It accessed the car's full state as well as the next 8m of the race line (20 waypoints, including suggested speed, multiplied by the tunable speed factor $F$) as illustrated in \cref{fig:f1tenth_car_track}. 

The cost function for \xx{nmpc} contains following terms (weights in brackets):
the quadratic difference between the car's position and the desired race line (20.0) %$d_r$
, the quadratic difference between the car's speed and the suggested speed of the previously calculated race line (0.1) % $d_y$
, the car's steering angle (30.0) % $\theta$
, the car's slipping angle (0.01) %$\beta$
, the first (0.4) and second derivative (10.0) of the steering history. %$d_\theta$and $dd_\theta$

 To compensate for the car's 80\,ms perception+actuation delay, rather than executing $\nu_k$ of the \xx{nmpc}'s optimal control sequence \cref{eq:1}, we executed $\nu_{k+4}$ (4/50Hz=80ms).  Like this, we execute the control that was calculated for the car's estimated state 80ms in the future.

\begin{figure}[thp]
  \centering
      \includegraphics[width=.99\columnwidth]{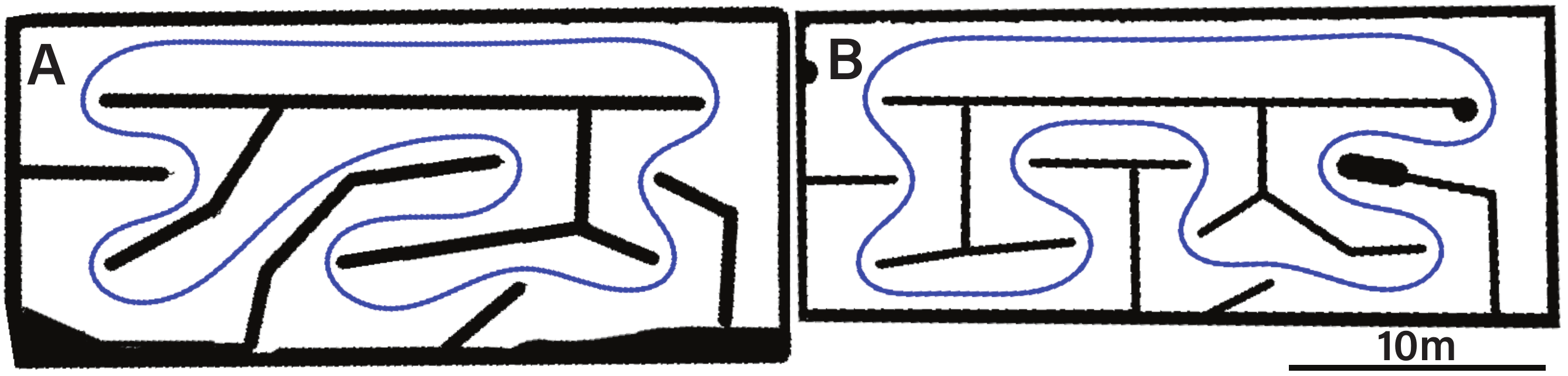}
      \caption{Maps recorded by the physical car with their minimum curvature race lines. \textbf{A:} RCA1, the map used for training, \textbf{B:} RCA2, the map used for experiments}
  \label{fig:f1tenth_maps}
\end{figure}

 We collected this training data in simulation by driving around the \cref{fig:f1tenth_maps}A track with the \xx{nmpc} for $600s$ for each speed factor $F \in [0.5, 0.6, 0.7, 0.8, 0.9, 1.0, 1.1, 1.2]$, so the training data consisted of $4,800s$, or $240,000$ samples.

The track layout (RCA1 in {\cref{fig:f1tenth_maps}A}) used for training is different than the one used for testing (RCA2 in {\cref{fig:f1tenth_maps}B}). This shows that the \xx{nc} can generalize over different tracks.

\subsubsection{\xx{nc} architecture}

\cref{tab:nn_params} lists the F1TENTH \xx{nc} parameters. In total, the \xx{nc} has 64 inputs and 2 outputs.
There are 2 hidden layers, each with 64 units.

The inputs of the car \xx{nc} consist of the next 20 waypoints relative to the car's position and heading, as described in \cref{subsubsec:f1tenth_data_collection} , the car's longitudinal velocity $v_x$, angular velocity $\omega_z$, steering angle $\theta$ and slipping angle $\beta$. The car's position and heading are used to convert the waypoints to the car's relative coordinate system but they are not part of the neural network input.

The \xx{nc} outputs are the desired longitudinal velocity $u_\text{vel}$ 
and the desired steering angle 
$u_\theta$ 
of the car.
The motor throttle is determined from 
$u_\text{vel}$ by the internal PID controller of the VESC.
The steering servo receives the desired angle $u_\theta$ and applies it with its internal PID controller on the steering wheels.

To train a functional \xx{nc}, we found it was essential to take into account the \xx{nmpc}'s delay compensation. We therefore also trained the \xx{nc} to output the future control step $\nu_{k+4}$ exactly like the \xx{nmpc}.

\subsubsection{Reference solutions}
\label{subsubsec:f1tenth_reference_solutions}

\begin{itemize}
\item{\textbf{\xx{nmpc}}}:
We used the \xx{nmpc} from \cref{subsubsec:f1tenth_data_collection} as the reference controller for optimal control.

\item{\textbf{\xx{pp} Controller}:}
As a simpler and more computationally efficient reference controller, we used an implementation of \xx{pp} \cite{snider2009PurePursuit}.
\xx{pp} is widely used in the F1TENTH competition due to its simplicity. The controller can be tuned easily, but it is not able to drive more complex maneuvers such as slipping or drifting.

In our work we use a \xx{pp} with a dynamic look-ahead distance $d_l$ proportional to the speed of the car with an empirically obtained proportionality factor (0.4). This formulation delivers a short $d_l$ (accurate tracking) in sharp curves and a larger $d_l$ (more stable) in high-speed sectors.

\end{itemize}
\section{EXPERIMENTAL RESULTS}
\label{sec:experimental_results}
This section presents the results of our experiments with the physical cartpole and the F1TENTH race car.

\subsection{Physical Cartpole Experiments}
\label{subsec:phyical_cartpole_experiments}

\begin{figure}[t]
  \centering
      \includegraphics[width=\columnwidth]{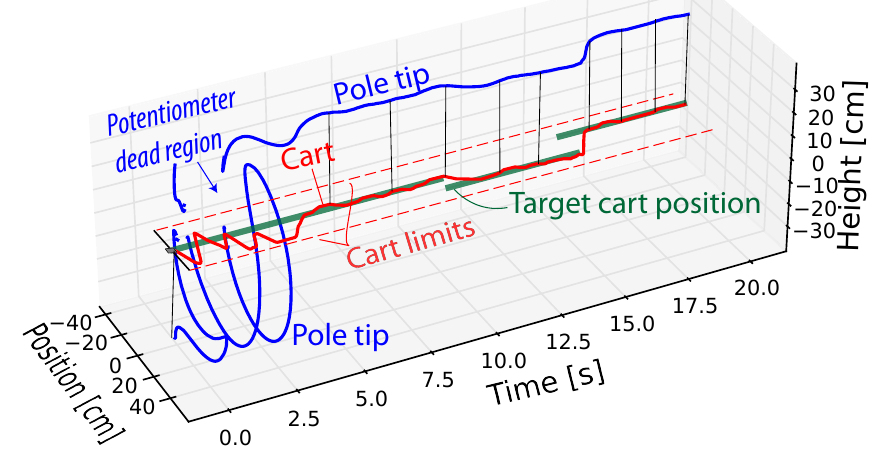}
      \caption{\textbf{Physical cartpole experiment.} The cart and pole tip positions are plotted over time for a swing-up followed by changes in the target cart position. The cartpole is controlled by the hardware \xx{nc}.}
  \label{fig:cartpole-3d}
\end{figure}

\cref{fig:cartpole-3d} shows a swing-up followed by changes in target cart position for the physical cartpole of \cref{subsec:methods_cartpole}. We can observe that the swing-up completes in about 2.5s. Changes in target cart position are accurately tracked.

\begin{figure}[thp]
  \centering
      \includegraphics[width=1.0\columnwidth]{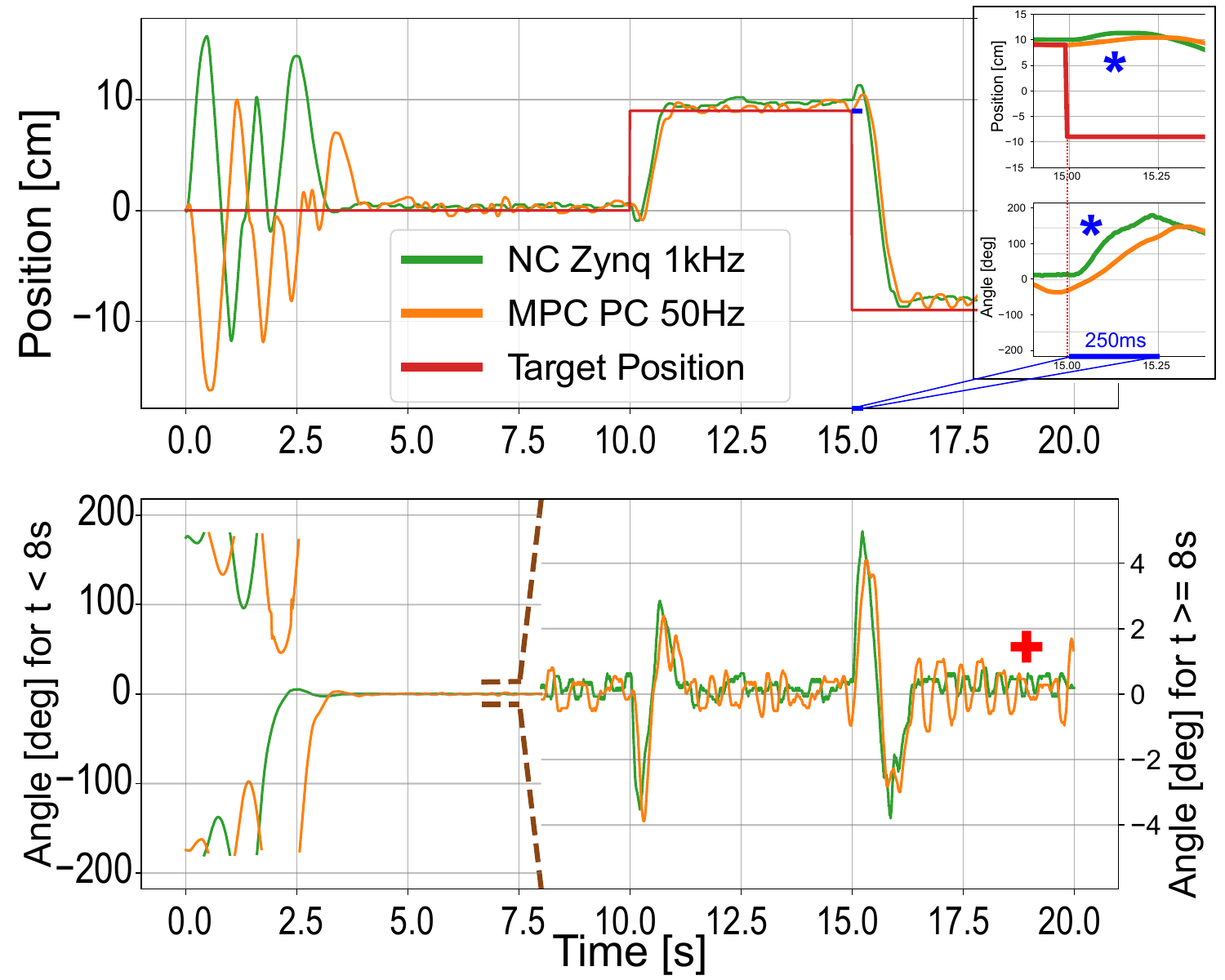}
      \caption{Comparison of cartpole dynamics with various controllers.}
  \label{fig:results_cartpole_comparison}
\end{figure}

\cref{fig:results_cartpole_comparison} compares the cartpole dynamics controlled by the hardware \xx{nc} at 1\,kHz and controlled by \xx{nmpc} running on a host computer at 50\,Hz. The controllers have similar control performance, however, in the inset (\textbf{{\textcolor{blue}*}}), we can see that the \xx{nc} reacts more quickly to the step change of target position.
The \xx{nc} pole error is also smaller ({\textcolor{red}+}).

\subsection{Simulated F1TENTH Car Experiments}
This section compares the \xx{nc} with the original \xx{nmpc}, and a \xx{pp} controller.
For all car experiments, the speed factor $F$ was experimentally adjusted as high as possible such that the car still could complete 10 laps in a row without crashing.

\begin{figure}[thp]
  \centering
      \includegraphics[width=.9\columnwidth]{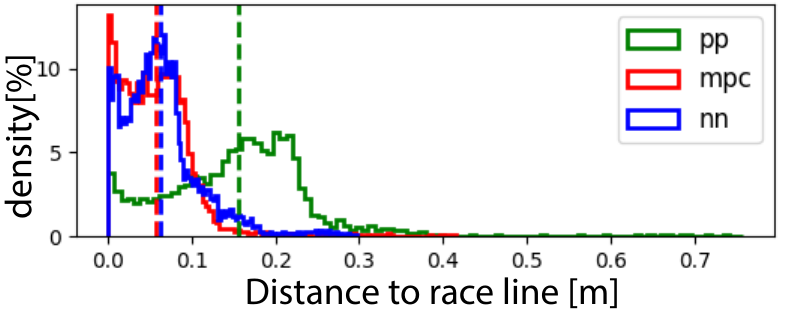}
      \caption{Histograms of distance errors to minimum curvature race line of all controllers, evaluated in the simulated environment
      See \cref{table-car-results-sim}.}
  \label{fig:sim_car_error_histograms}
\end{figure}

\cref{table-car-results-sim} summarizes the maximal 10-lap ($d_{max}$) and the mean ($d_{mean}$) distance to the race line as well as the achieved mean lap time ($t_{lap}$) of each controller. 
\cref{fig:sim_car_error_histograms} compares 10 simulated lap tracking error histograms from \xx{pp}, \xx{nmpc}, and the \xx{nc}. 
The \xx{nc} drove most precisely.
The \xx{pp} was 30\% slower than the \xx{nc}.
The \xx{nmpc} controller drove fastest and with smallest mean error, but the \xx{nc} had the smallest maximum error and was slower by only 10\% than the \xx{nmpc}.

\newcommand{\barwidth}{2} % cm max bar widths
\newcommand{\barheight}{2pt}
\newcommand{\laptimescale}{25} % max scale for percent bars
\newcommand{\degscale}{90} % max scale for degree bars

\def\laptime#1{%% percent bar
   {\color{red}\rule{\fpeval{#1/\laptimescale*\barwidth} cm}{\barheight}} #1
}
\def\laptimeb#1{%% best value
 {\color{blue}\rule{\fpeval{#1/\laptimescale*\barwidth} cm}{\barheight}} #1
}
\def\db#1{%% degrees bar
  {\color{red}\rule{\fpeval{#1/\degscale*\barwidth} cm}{\barheight}} #1
}
\def\dbb#1{%% best
  {\color{blue}\rule{\fpeval{#1/\degscale*\barwidth} cm}{\barheight}} #1
}

\begin{table}
\begin{center}
\caption{Simulation F1TENTH lap times and position accuracy.}
\label{table-car-results-sim}
\begin{tabular}{c c c c l} 
 Controller & $F$ & {{$d_\text{max}$} [m]} & {$d_\text{mean}$ [m]} & $t_\text{lap}$ [s] \\ [0.5ex] 
 \hline
\xx{pp} & 0.8 & 0.758 & 0.188 & \laptime{23.124} \\ 
{\xx{nmpc}}&\textbf{1.2} & 0.453 & \textbf{0.048} & \laptimeb{15.822} \\
\xx{nc} & 1.1& \textbf{0.317} & 0.086 & \laptime{17.51} \\
\end{tabular}
\end{center}
\end{table}

\subsection{Physical F1TENTH Car Experiments}
\label{subsec:phyical_f1tenth_experiments}
On the physical F1TENTH car, we compared the hardware \xx{nc} against a \xx{pp} controller. 
All results were measured with the state estimation stack described in \cref{subsubsec:physical_f1tenth_car}.

The \xx{nmpc} was not able to complete laps with the physical car due to the lack of onboard computational power: The \xx{nmpc} ran at a maximum of 10Hz, but only by consuming all the computational resources needed by the state estimation stack.

\cref{table-car-results-physical} compares the the performance metrics as in \cref{table-car-results-sim} over 10 laps.

\begin{figure}[thp]
  \centering
      \includegraphics[width=.8\columnwidth]{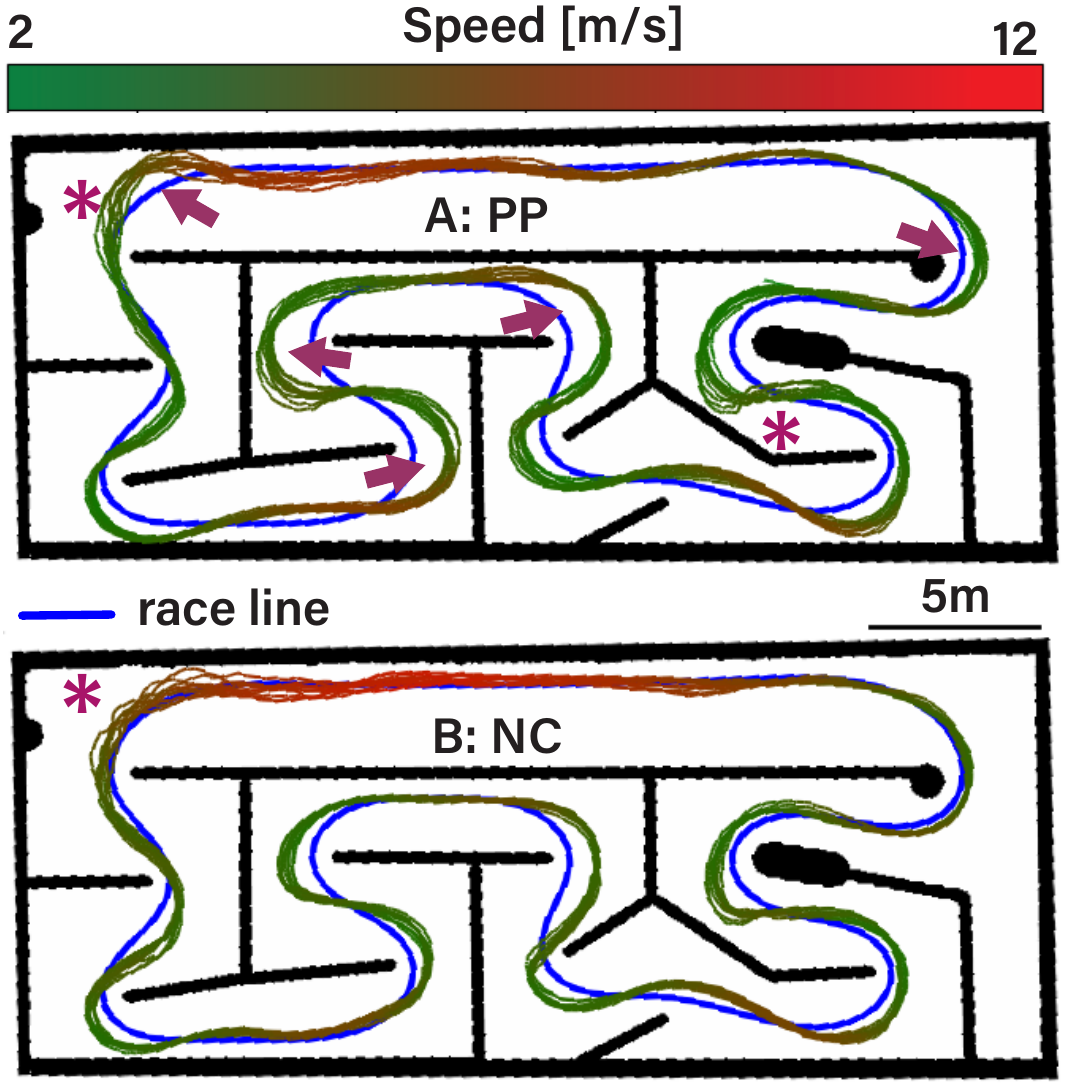}
      \caption{Trajectories of the physical car controlled with \textbf{A:} Pure Pursuit and \textbf{B:} \xx{nc} following minimum curvature race line.  See \cref{table-car-results-physical}.}
  \label{fig:physical_pp_trajectories}
\end{figure}

\cref{fig:physical_pp_trajectories} compares 10 lap trajectories of the physical car controlled by the \xx{pp} controller and the \xx{nc}. 
The arrows 
(\textcolor{BrickRed}{$\boldsymbol{\rightarrow}$}) 
show regions where our \xx{nc} noticeably outperforms the reference \xx{pp} in terms of accuracy. 
We suspect that in regions marked by 
(\textbf{{\textcolor{BrickRed}*}}),
 the recorded trajectory deviates noticeably from the ground truth. The state estimation stack might be inaccurate at high speed or high angular velocity.

\begin{figure}[htb]
  \centering
      \includegraphics[width=.8\columnwidth]{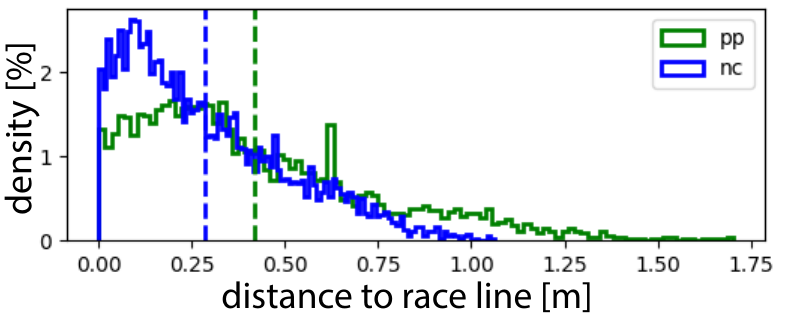}
      \caption{Histograms of distance errors to minimum curvature race line of the \xx{pp} and \xx{nc} controllers on the physical car.  See \cref{table-car-results-physical}.}
  \label{fig:physical_tracking_errors}
\end{figure}

\cref{fig:physical_tracking_errors} illustrates the according tracking error histograms from \xx{pp} and the \xx{nc}.

The \xx{nc} lap times are 20\% quicker than the \xx{pp} times. The \xx{nc} tracks the race line more precise than the \xx{pp}. The \xx{pp} overshoots the optimal line more than the \xx{nc}, particularly on sharp curves.

\begin{table}
\begin{center}
\caption{Physical F1TENTH car lap times and position accuracy.}
\label{table-car-results-physical}
\begin{tabular}{c c c c l} 
 Controller & $F$ & $d_\text{max}$[m] & $d_\text{mean}$[m] & $t_\text{lap}$ [s] \\ [0.5ex] 
 \hline
\xx{pp}& 0.71  & 1.80 & 0.453 & \laptime{24.35} \\ 
\xx{nc} & \textbf{0.86} & \textbf{1.098} & \textbf{0.326} & \laptimeb{20.04} \\
\end{tabular}
\end{center}
\end{table}

\section{CONCLUSION}
\label{sec:conclusion}
For the cartpole robot, we showed that the hardware \xx{nc} achieves control performance that matches the expensive \xx{nmpc}. The \xx{soc} \xx{fpga} forms the complete controller of the system including the sensor and the motor interface. The hardware \xx{nc} enables a control rate of 1\,kHz, which through quick corrections helps reducing angle deviation around equilibrium.

In the F1TENTH car simulation, we showed that the hardware \xx{nc} follows the race line more accurately and achieves faster lap times than the reference pure pursuit controller. We showed that it competes with its teacher \xx{nmpc} in terms of accuracy and lap times with a small fraction of the computational effort and latency. The \xx{nc} can generalize and complete laps on a track with a different distribution of curves than the one it was trained on.

In the physical car, we demonstrated a real-world application that offloads all the control computation onto a low-power and low-cost \xx{soc} \xx{fpga}. While the \xx{nmpc} can not be solved in real time on the car's computer, the hardware \xx{nc} is able to imitate it and therefore outperforms (in accuracy and lap times) the reference pure pursuit controller.

\subsection{Limitations and Outlook}
Like \xx{nmpc}, our \xxx{nc} control performance is degraded by Sim2Real model mismatch. 
This limitation could be addressed by training with domain randomization as is done in \xx{rl}~\cite{Song2023-mpc-vs-rl-drone-racing}. 

\xxx{nc} provide a framework to infer state components and environmental variables that are not directly accessible during controller operation. An example is \xx{rma} in \cite{Kumar2021-rma-legged-robots} or the teacher-student \xx{rl} method used in~\cite{Lee2020-rsl-quadroped-locomotion-rl}. In the same way, if we provide sufficient temporal context, it may be possible to train the \xxx{nc} for our robots with variable physical properties, such as pole lengths or track frictions. These and similar methods might be useful for car control in the absence of slip angle information which is difficult to measure.
This would show further advantages of \xxx{nc}.

The hardware \xxx{nc} are not restricted to imitate \xx{nmpc} controllers with the small \xxx{mlp} we needed for this work and can easily be adapted to run large \xxx{rnn}~\cite{Gao2020-edgedrnn} or
 \xxx{mlp}~\cite{Chen2021-eile-mlp-training-accelerator} 
trained by other methods.

%%%%%%%%%%%%%%%%%%%%%%%%%%%%%%%%%%%%%%%%%%%%%%%%%%%%%%%%%%%%%%%%%%%%%%%%%%%%%%%%
\footnotesize
{
\section*{ACKNOWLEDGMENTS}
This work was supported by the Swiss NSF projects SCIDVS (200021\_185069) and VIPS (40B2-0\_181010) and by the Samsung Global Research Neuromorphic Processor Project (NPP).
% (\url{https://sensors.ini.ch/research/areas/neural-hardware}).
We thank Daniel Schüpbach for allowing us to use his indoor track in the RC Arena in Gachnang, Switzerland.
We thank our students Jago Iranyi, Nigalsan Ravichandran and Eike Himstedt for exploring the topics of this work with us.
We thank Pehuen Moure, Ilya Kiselev, and Shih-Chii Liu for helpful comments and the 2023 Telluride Neuromorphic Workshop (\url{https://tellurideneuromorphic.org}) for the opportunity to develop this work.  
}

\clearpage
\newpage
\newpage

\AtNextBibliography{\footnotesize}{
\printbibliography
}
% % \AtNextBibliography{\footnotesize}
% \bibliography{bibliography}

\end{document}

\typeout{get arXiv to do 4 passes: Label(s) may have changed. Rerun}